\documentclass{article}
\usepackage{amsmath,epsfig}
\usepackage[preprint]{spconfa4}
\usepackage[misc]{ifsym}
\usepackage{algorithm}
\usepackage{bm}
\usepackage{xcolor}
\usepackage{amssymb}
\usepackage{multirow}
\usepackage{cite}
\usepackage{hyperref}

\let\OLDthebibliography\thebibliography
\renewcommand\thebibliography[1]{
  \OLDthebibliography{#1}
  \setlength{\parskip}{0pt}
  \setlength{\itemsep}{0pt plus 0.3ex}
}

\begin{document}\sloppy

\def\x{{\mathbf x}}
\def\L{{\cal L}}

\title{Continual Contrastive Learning for Image Classification}
%
\name{Zhiwei Lin, Yongtao Wang$^{\textrm{\Letter}}$, Hongxiang Lin }
\address{Wangxuan Institute of Computer Technology, Peking University, Beijing, China\\
\{zwlin, wyt\}@pku.edu.cn
}

\maketitle

\begin{abstract}
Recently, self-supervised representation learning gives further development in multimedia technology. Most existing self-supervised learning methods are applicable to packaged data. However, when it comes to streamed data, they are suffering from catastrophic forgetting problem, which is not studied extensively. 
In this paper, we make the first attempt to tackle the catastrophic forgetting problem in the mainstream self-supervised methods, \textit{i.e.}, contrastive learning methods. Specifically, we first develop a rehearsal-based framework combined with a novel sampling strategy and a self-supervised knowledge distillation to transfer information over time efficiently. Then, we propose an extra sample queue to help the network separate the feature representations of old and new data in the embedding space.
Experimental results show that compared with the naive self-supervised baseline, which learns tasks one by one without taking any technique, we improve the image classification accuracy by $1.60\%$ on CIFAR-100, $2.86\%$ on ImageNet-Sub, and $1.29\%$ on ImageNet-Full under 10 incremental steps setting.
Our code will be available at \href{https://github.com/VDIGPKU/ContinualContrastiveLearning}{https://github.com/VDIGPKU/ContinualContrastiveLearning}.


\end{abstract}
\section{Introduction}
\label{sec:intro}
Recently, as a novel branch of unsupervised learning, self-supervised learning is proposed to learn general feature representations from unlabeled data, and further boosts the performance of multimedia technology. The great success of self-supervised learning on visual representation relies on its impressive potential of learning from unlabeled data on a large scale. Many experiments show that, with the larger and more diverse training data, the self-supervised model can learn a better feature representation. However, in most practical scenarios, the unlabeled training data are streamed. 
The limitation of storage and computational power does not allow us to collect all data together and use it to train the model.\par

\begin{figure}[t]

\begin{minipage}[b]{.48\linewidth}
  \centering
\centerline{\epsfig{figure=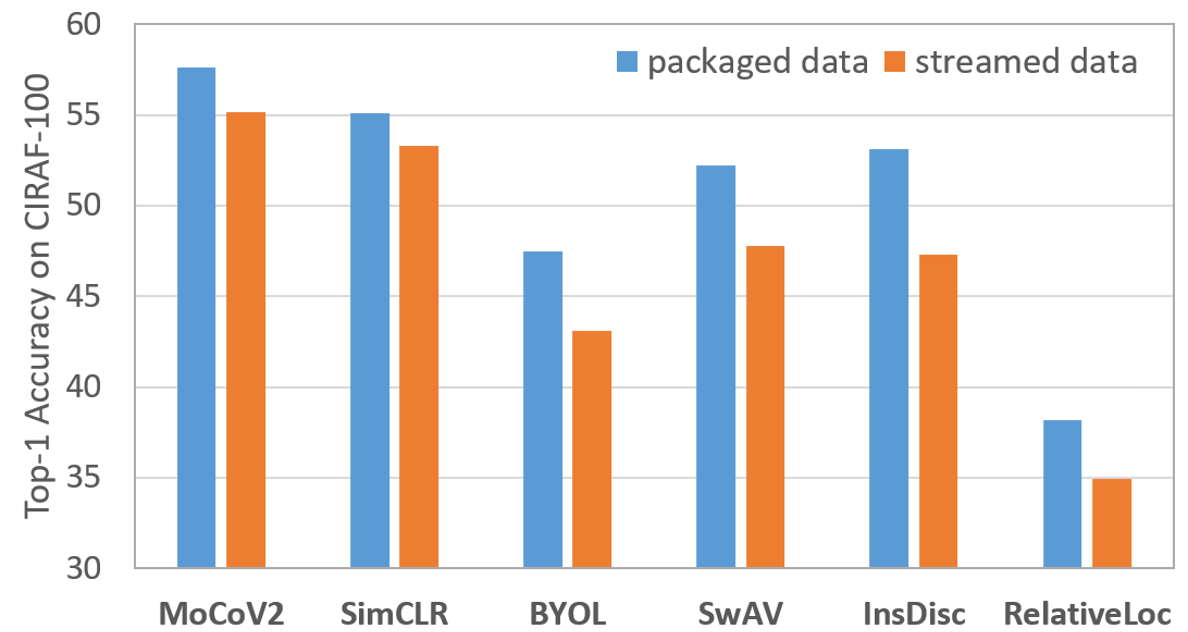,width=4.3cm}}
  \centerline{(a)}\medskip
\end{minipage}
\hfill
\begin{minipage}[b]{0.48\linewidth}
  \centering
\centerline{\epsfig{figure=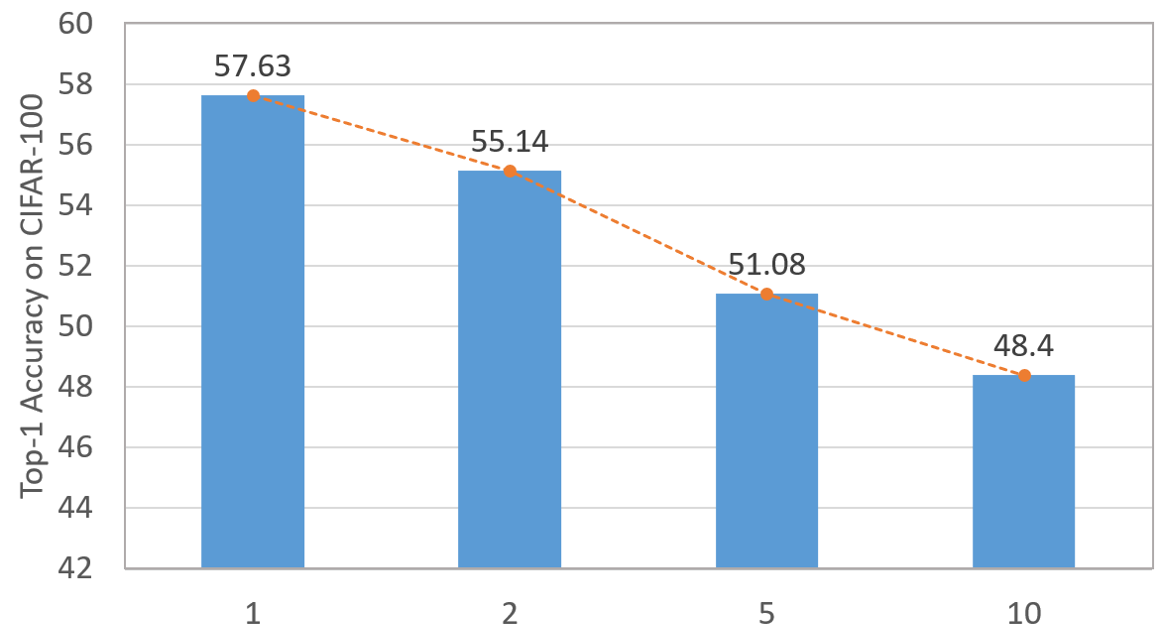,width=4.0cm}}
  \centerline{(b)}\medskip
\end{minipage}
%
	\caption{Illustration of the catastrophic forgetting in self-supervised learning. (a) The performance degradations of several self-supervised learning methods on CIFAR-100; (b) Linear evaluation top-1 accuracy on CIFAR-100 of MoCoV2 under different Class-Incremental Learning settings.}
\label{fig:1}
\end{figure}

For streamed data, directly training the model on it and updating its parameters causes the catastrophic forgetting problem, namely, a drastic performance drop on the old data when the model is trained on new data. The catastrophic forgetting problem has been discovered and studied in supervised learning on many tasks~\cite{icarl}\cite{ob2}\cite{plop}. However, under the unsupervised learning setting, the problem is seldom studied~\cite{cuper}\cite{wild}. In our experiments, we find that self-supervised learning is also suffered from catastrophic forgetting. For example, Fig. \ref{fig:1}(a) illustrates that the catastrophic forgetting in several self-supervised learning methods. Obviously, the classification performance of these methods decreases when they learn the streamed data. Moreover, Fig. \ref{fig:1}(b) shows that when the number of incremental steps increases from 2 to 10, MoCoV2~\cite{mocov2} forgets more knowledge it has learned. Hence, a continual self-supervised learning method is required.

In this paper, we try to alleviate the catastrophic forgetting in self-supervised learning. Specifically, we focus on the mainstream of self-supervised learning, contrastive learning, and consider a specific continual learning task, \textit{i.e.}, Class-Incremental Learning, in which each incremental data consists of new class samples. 
We restore parts of samples from old data in each incremental step and develop a rehearsal-based framework. In detail, first, to select more representative samples, we propose a novel sampling strategy, which is different from the previous sampling strategy in supervised continual learning in the following aspects: (1) our sampling strategy does not rely on the data labels; (2) the samples are sorted by the feature variance instead of the confidence of classification. Second, to utilize the stored samples efficiently, we further introduce self-supervised knowledge distillation to our framework, to better transfer the feature representation learned before. Moreover, though the properties of feature representation of old data are sustained by samples selection and knowledge distillation, when the network learns feature representations of new data, the region of feature vectors of new data in embedding space may mix up with the region of feature vectors of old data (as illustrated in Fig. \ref{fig:i1}). To address this issue, we use an extra sample queue to help the network discriminate new data from old data.

The main contributions of this work can be summarized as:
\begin{itemize}
	\item To the best of our knowledge, we are the first to address the problem of catastrophic forgetting in contrastive learning. 
	\item We develop a rehearsal-based framework, which utilizes a novel sampling strategy and self-supervised knowledge distillation to transfer knowledge from old data.
	\item We propose an extra sample queue to reduce the interference of feature representation distribution when the network learns a new task.
	\item We improve the performance of our baseline methods on the CIFAR-100 and ImageNet significantly.
\end{itemize}

\section{Related Work}
\subsection{Self-supervised Learning}
The core idea of self-supervised learning is to train a neural network on the large-scale unlabeled data by designing specific pretext tasks, so that the network can learn a general and transferable feature representation. At present, 
instance discrimination~\cite{insdisc}, or more specifically, contrastive learning~\cite{moco}\cite{simclr}\cite{simsiam}\cite{byol}, has made breakthroughs and achieves new state-of-the-art performance on feature representation learning. In instance discrimination methods, one image falls into one category by distinguishing from other images, and the learning objective is to maximize the mutual information of two views from the same image generated by different data augmentation operations.\par

When training a self-supervised learning network, a large-scale data set is indispensable. However, directly training a whole large data set may be unaffordable in many scenarios. Moreover, new data is usually generated continuously. Collecting old and new data together and training the network again is a time and resources consuming solution. Thus, a continuous style of self-supervised learning method is imperative when it comes to application.

\subsection{Class Incremental Learning}
Class-Incremental Learning (CIL) aims to learn a classification model with the number of classes increasing step by step. Existing works on CIL often adopt rehearsal methods and knowledge distillation to tackle the catastrophic forgetting problem. Rehearsal-based methods~\cite{icarl}\cite{LU} try to select a representative set of samples from the old data. Specifically, these works use data labels or use a classifier trained with labeled data, to select the samples. However, in self-supervised continual learning, there is no access to the data labels. Thus, in our method, we propose a novel sampling strategy using feature variance measurement, which does not rely on data labels. Distillation-based methods~\cite{lwf}\cite{ssil} use knowledge distillation loss as the regularization term to preserve previous knowledge when learning new data. Many distillation-based methods use supervised knowledge distillation, \textit{i.e.}, the predicted label logits of the new model are enforced to be close to those of the old model. However, directly applying supervised knowledge distillation loss to self-supervised continual learning will cause some problems since it lacks the classification predictor. Thus, we adopt self-supervised knowledge distillation loss~\cite{seed} in our method, which does not require the classification head. Moreover, unlike traditional knowledge distillation methods which often fix the teacher network during distillation, we propose the momentum teacher design.

\section{Proposed Method}
\begin{figure}
	\centering
    \includegraphics[width=1.0\linewidth]{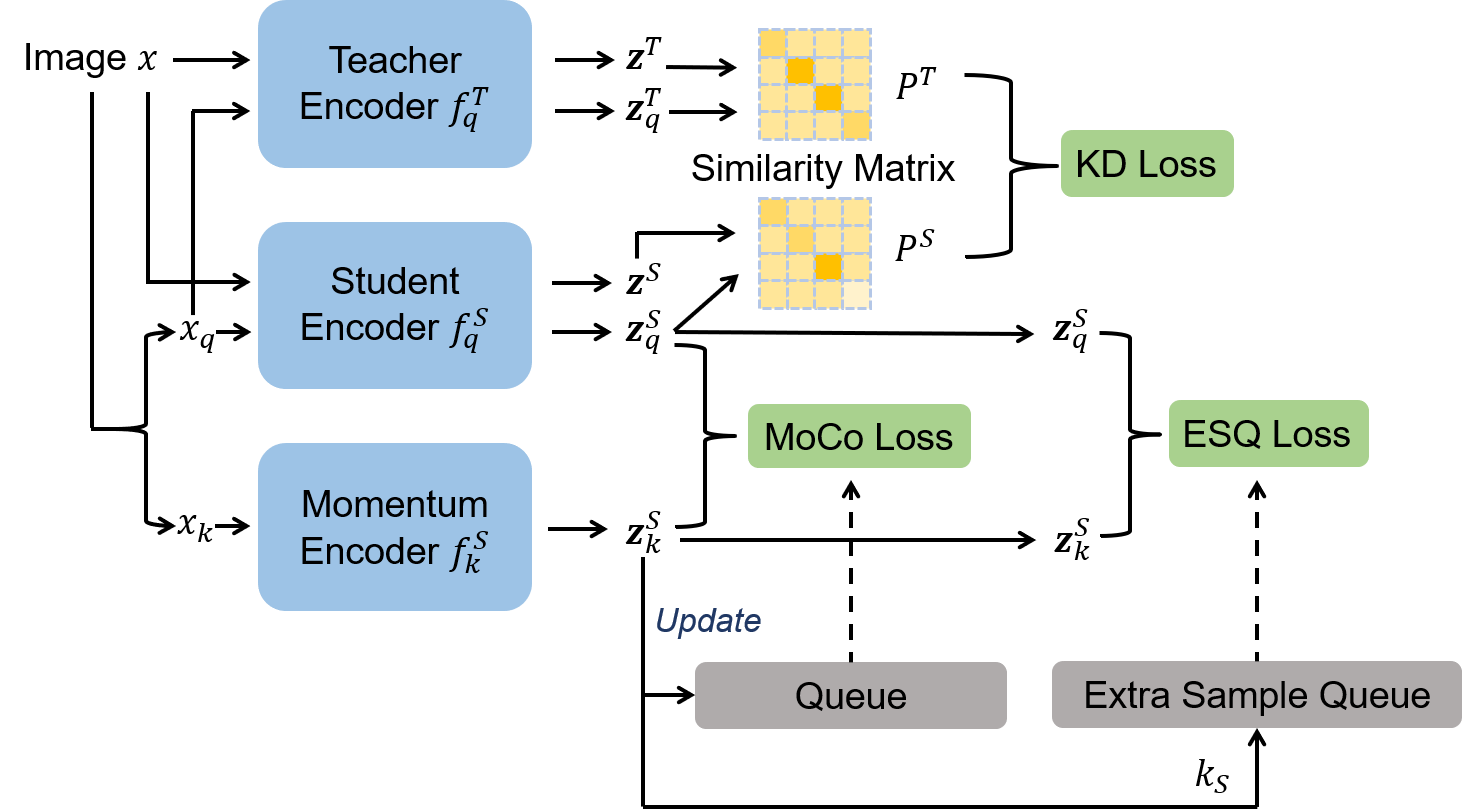}
	\caption{The pipeline of our Continual Contrastive Learning method.}
	\label{fig:2}
\end{figure}
We implement our Continual Contrastive Learning (CCL) method based on the widely used contrastive learning framework MoCoV2~\cite{mocov2}, and the overall pipeline is shown in Fig. \ref{fig:2}.The main components of our method are introduced as following.\par

\subsection{MoCoV2} \label{section:moco}
First, we introduce MoCoV2~\cite{mocov2} briefly for better understanding. MoCoV2~\cite{mocov2} contains two encoders, $f_q$ and $f_k$, and a memory bank. Given an input image $x$, we first transform it to two views $x_q$ and $x_k$ with two different augmentations. Then, we get query vector $q$ and its positive key sample $k_+$ by feeding $x_q$, $x_k$ into $f_q$ and $f_k$ respectively. The memory bank stores negative key samples $Q=\{{k_1},{k_2}, {k_3}, ..., {k_n}\}$ of $q$. Finally, MoCoV2~\cite{mocov2} adopts a contrastive loss to pull $q$ to $k_+$ and push away $q$ from $Q=\{{k_1},{k_2}, {k_3}, ..., {k_n}\}$.\par
Specifically, the contrastive loss is defined as:
\begin{equation}
\mathcal{L}_{contrast}(q, k_+, Q)=-\log \frac{\exp(q\cdot k_+/\tau)}{\sum_{k_i\in Q\cup\{k_+\}}\exp(q\cdot k_i/\tau)},
\end{equation}
where $\tau$ is the hyper-parameter of temperature and $n$ is the number of negative samples.\par
Moreover, two encoders $f_q$ and $f_k$ share the same architecture, \textit{i.e.}, the backbone of the network followed by an extra MLP, but with different weights $\theta_q$ and $\theta_k$. The parameters $\theta_q$ are updated by back-propagation and the parameters $\theta_k$ are updated by momentum strategy, that is,
\begin{equation}
\theta_k \leftarrow m\theta_k+(1-m)\theta_q.
\end{equation}

\subsection{Rehearsal with Knowledge Distillation}\label{section:rehearsal}
Rehearsal is widely used in supervised continual learning. By replaying the stored old samples from previous tasks, we find rehearsal methods can also help alleviate catastrophe forgetting in self-supervised continual learning. Besides, inspired by many supervised continual learning methods, we utilize self-supervised knowledge distillation to transfer the contrastive information when the network learns a new task.\par 

In continual learning, given a set of datasets $\{D_1, D_2, ..., D_t, ..., D_N\}$, the network will be trained continuously on the dataset $D_t$ at time t. In rehearsal methods, we store a small fraction of data from $D_t$ after the training of $D_t$ is finished. Instead of randomly storing samples from old data, we propose a novel sampling strategy with the feature variance measurement. Specifically, when the training on $D_{t}$ is finished, we input all images of $D_{t}$ into the encoder $f_q$ to extract feature vectors. Then we group the vectors by the K-Means algorithm and divide them into $C$ classes. In our experiments, we set the $C$ equal to the number of classes in $D_{t}$. Then, for images $x$ in class $c$, we have different views $\{\mathcal{T}_1(x),\mathcal{T}_2(x),...,\mathcal{T}_l(x)\}$ of $x$ after applying different data augmentations $\mathcal{T}$ mentioned in the MoCoV2~\cite{mocov2}. After that, we calculate the variance of the feature vectors of these views $\{f_q(\mathcal{T}_1(x)),f_q(\mathcal{T}_2(x)),$ $ ..., f_q(\mathcal{T}_l(x))\}$. We found $l=6$ is enough for measuring the variance. Finally, we store the first $n$ images with the smallest variance for each class. Essentially, the images with the smallest variance are ones well learned by the network, thus the contrastive information of these images can be easily transferred to the new task through self-supervised knowledge distillation.\par

Knowledge distillation is another way to address catastrophe forgetting problems. Hence, we introduce it to our method to further enhance the ability for handling catastrophe forgetting. Following some self-supervised knowledge distillation methods~\cite{seed}\cite{meet}, we utilize the contrastive similarity between images from the teacher network as the distillation target for the student network.
Specifically, given the sampled images $x$ from the old dataset, we have $x_q$ after one data augmentation. Then, $x$ and $x_q$ are mapped and normalized into feature vector representations $\textbf{z}^T=f_{q}^T(x)$, $\textbf{z}^T_q=f_{q}^T(x_q)$,  $\textbf{z}^S=f_{q}^S(x)$ and $\textbf{z}^S_q=f_{q}^S(x_q) \in \mathbb{R}^{B\times D}$, where $B$ is the number of images, $D$ is the feature dimension, and $f_{q}^T$ and $f_{q}^S$ denote the query encoder of teacher and student respectively. We compute the similarity matrix $P^T(\textbf{z}^T, \textbf{z}^T_{q}) \in \mathbb{R}^{B\times B}$ between the feature vectors $\textbf{z}^T$ and $\textbf{z}^T_{q}$ extracted by teacher.
After that, the softmax function with the temperature scaling factor $\tau$ is applied to $P^T$ for normalization. We obtain $P^S$ with a similar operation from the student. Finally, we adopt KL-divergence loss to $P^T$ and $P^S$:

\begin{equation}
\mathcal{L}_{kd}=-\sum_{i=1}^b\sum_{j=1}^b P^T_{ij}\log(P^S_{ij}).
\end{equation}
\par
Besides, compared to supervised continual learning, which often fixes the teacher network during distillation, we newly find that a momentum teacher brings better performance. Thus, for each training epoch, we update the teacher network with a momentum style:
\begin{equation}
\theta_t \leftarrow m_t\theta_t+(1-m_t)\theta_q,
\end{equation}
where $m_t=0.996$ in our experiments.\par

\begin{figure}
	\centering
	\includegraphics[width=1.0\linewidth]{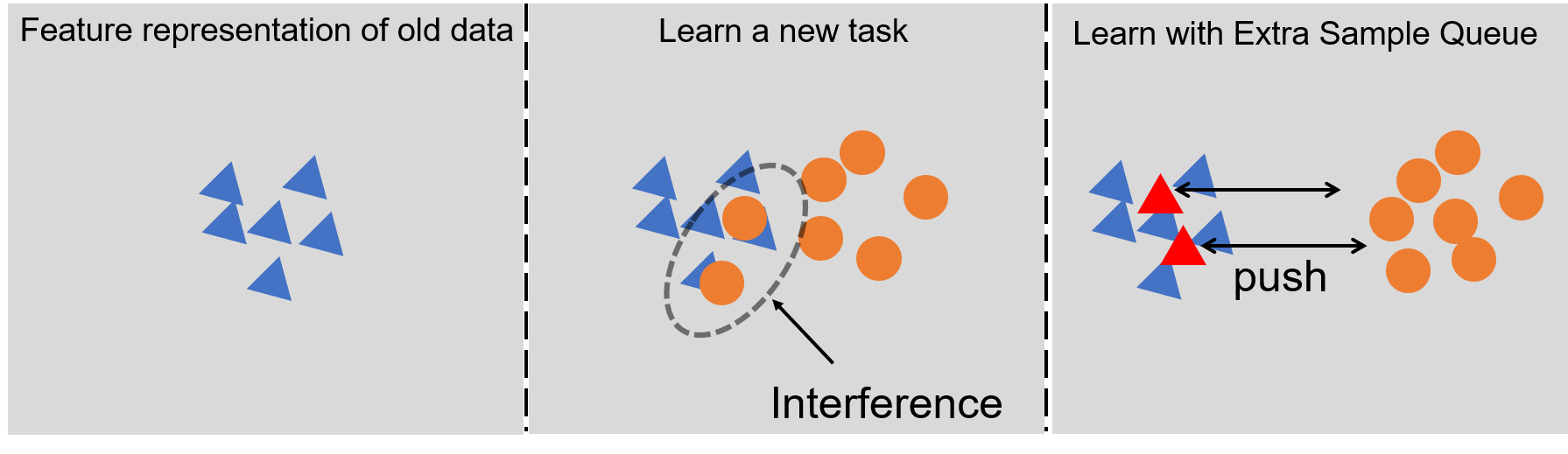}
	\caption{The illustration of feature space interference of old and new data and the effect of the extra sample queue. The blue triangles denote the feature vectors of old data. The orange circulars indicate the feature vectors of new data. The red triangles denote the negative samples stored in the extra sample queue.}
	\label{fig:i1}
\end{figure}

\subsection{Extra Sample Queue}\label{section:queue}
As mentioned in the section of the introduction, learning feature representations of a new task will cause the mixture of feature space among the old data and the new data. As illustrated in Fig. \ref{fig:i1}, when the network learns the feature representation (orange circulars) of new data, it may interfere with the feature representation (blue triangles) of old data.
Thus, we propose an extra sample queue (red triangles) to push away the feature of new data from the feature of old data.\par
Specifically, the extra sample queue only contains the negative samples of the sampled images from old data. For a data mini-batch $B=B_S\cup B_D$, where $B_S$ is the sampled images from old data and $B_D$ is the images from new data, we sent them to two encoders of MoCoV2~\cite{mocov2} and then obtain feature representations $\textbf{z}^S_q=q_S\cup q_D$ and $\textbf{z}^S_k=k_S\cup k_D$. The network discriminates data of the new dataset from data of the old datasets by computing the contrastive loss $\mathcal{L}_{ESQ}$ between $\textbf{z}^S_q$, $\textbf{z}^S_k$, and extra sample queue. In every iteration, we will pick out the feature $k_S$, and update the extra sample queue with $k_S$.\par 

To sum up, the total loss is given by:
\begin{equation}
\mathcal{L} = \lambda_1 \mathcal{L}_{MoCo} + \lambda_2 \mathcal{L}_{ESQ}+\lambda_3 \mathcal{L}_{kd},
\end{equation}
where $\lambda_1$, $\lambda_2$, and $\lambda_3$ are balancing weights. In our experiments, we set $\lambda_1=0.9$, $\lambda_2=0.1$ and $\lambda_3=0.1$. 

\begin{table*}[!t]
	\begin{center}
	\caption{
	Linear evaluation top-1 accuracy on CIFAR-100 and ImageNet-Sub\&Full under different incremental steps settings.
}
    \label{tab:CIFAR-100}
	\begin{tabular}{|c|c|c|c|c|c|}
		\hline  
		\multirow{2}{*}{Method}&\multicolumn{2}{c|}{\textbf{CIFAR-100}}&\multicolumn{2}{c|}{\textbf{ImageNet-Sub}}&\textbf{ImageNet-Full}\\

    \cline{2-6}
    & T=5 & 10 & 5 & 10 & 10 \\

		 \hline
		 Upper Bound& 57.63~$\pm$~0.12 & 57.63~$\pm$~0.12 & 60.54&60.54&67.50\\
		 Finetuning & 51.08~$\pm$~0.34& 48.40~$\pm$~0.42&56.56&52.62&61.50\\
		 Simple Rehearsal & 52.61~$\pm$~0.41 & 49.44~$\pm$~0.36&57.98&54.28&62.05 \\
		 Our & $\textbf{53.80}\pm0.37$ & $\textbf{50.10}\pm0.35$ &\textbf{58.92}&\textbf{55.48}&\textbf{62.79}\\

		\hline  
		\end{tabular}
	\end{center}
\end{table*}

\section{Experiment}
\subsection{Settings}
\subsubsection{Datasets.} We evaluate our method on two popular datasets for class-incremental learning, \textit{i.e.}, CIFAR-100 and ImageNet-Sub\&Full. Specifically, ImageNet-Sub is a subset of ImageNet-Full, which contains 100 classes of images selected from ImageNet-Full randomly. 
For a given dataset, the classes are first arranged in a fixed random order and then cut into T incremental splits that come sequentially.

\subsubsection{Protocols.} We follow~\cite{unbalance} to split the dataset into incremental steps and manage the memory budget. 1) For each split dataset, all classes are divided equally to come in T incremental steps. 2) For the old training set, a constant number of images are stored after each incremental step.\par

\subsubsection{Evaluation Metrics.} All models are evaluated after the last incremental step. To evaluate the performance of the encoders, we follow previous self-supervised learning tasks~\cite{insdisc}\cite{moco}\cite{simclr}, that is, we verify the encoders by linear classification on frozen features in term of top-1 accuracy.\par

\subsubsection{Implementation Details.} The experiments are conducted on CIFAR-100 with a modified ResNet18 as the backbone and on ImageNet-Sub\&Full with ResNet50 as the backbone. We adopt MoCoV2~\cite{mocov2} as our basic method. Each incremental training step consists of 200 epochs. For ImageNet-Sub\&Full, we use the same training hyperparameters as MoCoV2~\cite{mocov2}. For CIFAR-100, we follow the training hyperparameters in the official implementation of MoCoV2~\cite{mocov2} on CIFAR-10, and Split BN is used to simulate 8 GPU behavior of BatchNorm in 1 GPU. The size of the extra sample queue is set to 128 for both ImageNet-Sub\&Full and CIFAR-100. For linear classification, we follow the setting of MoCoV2~\cite{mocov2}. For each dataset, We divide the classes of the dataset into T parts equally with a random order. We store 20 images for each class in every incremental step. \emph{The experiment results on CIFAR-100 are obtained from the average of 6 trials}.

\subsection{Main Results}
Besides our method, we implemented and tested two naive continual contrastive learning methods, \emph{Finetuning} and \emph{Simple Rehearsal}. \emph{Finetuning} learns new tasks one by one without taking any technique to prevent catastrophic forgetting. \emph{Simple Rehearsal} stores a constant number of images randomly after training on the dataset $D_{t-1}$. For the subsequent task $t$, it adds the stored images into the new training set $D_t$ to train the model. It's worth noting that, \emph{Simple Rehearsal} does not use techniques like knowledge distillation or the extra sample queue.\par 

Table \ref{tab:CIFAR-100} shows the performance of different continual contrastive learning methods on CIFAR-100 and ImageNet. One can see that our method outperforms the others from each column of Table \ref{tab:CIFAR-100}.
More specifically, the top-1 accuracy is improved from $48.40\%$ to $50.10\%$ on CIFAR-100, from $52.62\%$ to $55.48\%$ on ImageNet-Sub, and from $61.50\%$ to $62.79\%$ on ImageNet-Full under 10 incremental settings. These results show that our method works well on both small and large datasets, and demonstrate the effectiveness of our proposed continual contrastive learning method.\par

Besides MoCoV2~\cite{mocov2}, we also apply our method to other contrastive learning methods, including SimCLR~\cite{simclr} and InsDisc~\cite{insdisc}. More detail can be found in Appendix. 
The results on CIFAR-100 are shown in Fig. \ref{fig:gen}. We can see that our method also improves the linear classification accuracy of other contrastive learning methods, which demonstrates the generalization of our method.

\begin{figure}
	\centering
	\includegraphics[width=1.0\linewidth]{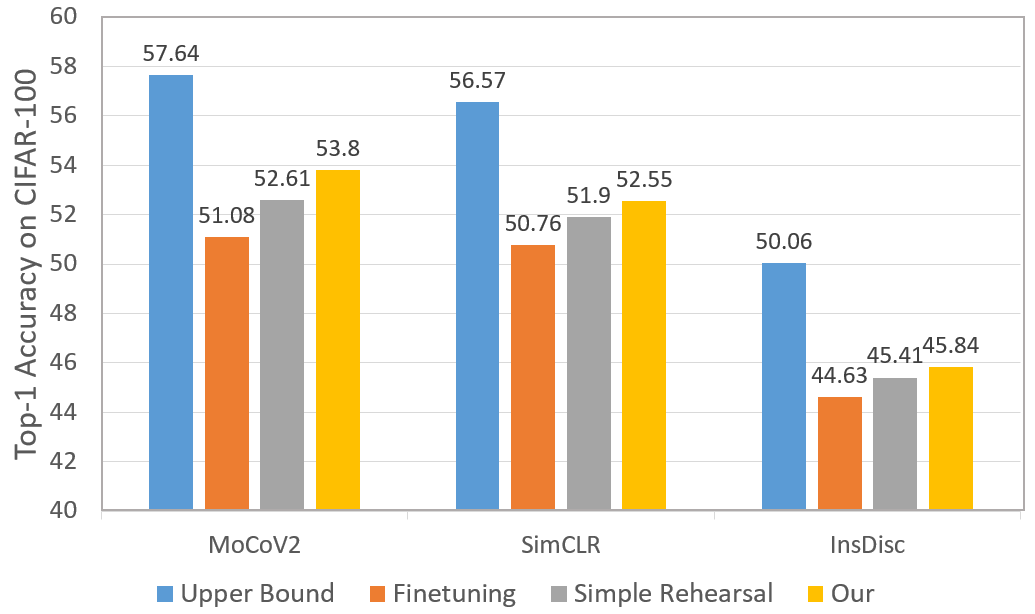}
	\caption{Linear evaluation top-1 accuracy of three contrastive learning methods on CIFAR-100 under 5 incremental steps.}
	\label{fig:gen}
\end{figure}

\begin{table}[t]
	\begin{center}
	\caption{\label{tab:kd and esq}
    Ablation studies of the main components of our method. The numbers in the brackets are the performance improvement compared with Finetuning. 
}

    \begin{tabular}{|c|c|}
		\hline  
		 Method & Top-1 Acc \\ 
		\hline  
		 Finetuning& 52.62 \\
		 +~Random Sampling& 54.28~(+1.66) \\
		 +~Our Sampling Strategy & 55.18~(+2.56)\\
		 +~Knowledge Distillation& 55.21~(+2.60) \\
		 +~Extra Sample Queue & \textbf{55.48~(+2.86)} \\
		\hline  
	\end{tabular}
\end{center}
\end{table}

\subsection{Ablation Study} 
To test the effectiveness of each component in our approach, we conduct ablation studies on ImageNet-Sub under 10 incremental steps setting.\par 

In the first experiment, we evaluate the main components of our method: sampling strategy with feature variance measurement, self-supervised knowledge distillation, and the extra sample queue. As described in Section \ref{section:rehearsal} and \ref{section:queue}, data sampling strategy and knowledge distillation are exploited to distill the contrastive information from the previous contrast process, and the extra sample queue is exploited to separate the feature space of the new data and old data. As shown in Table \ref{tab:kd and esq}, our novel sampling strategy, knowledge distillation, and extra sample queue boost the performance consistently and gains $2.86\%$ accuracy improvement when all of them are applied.\par

\begin{figure}[t]
	\centering
	\includegraphics[width=0.9\linewidth]{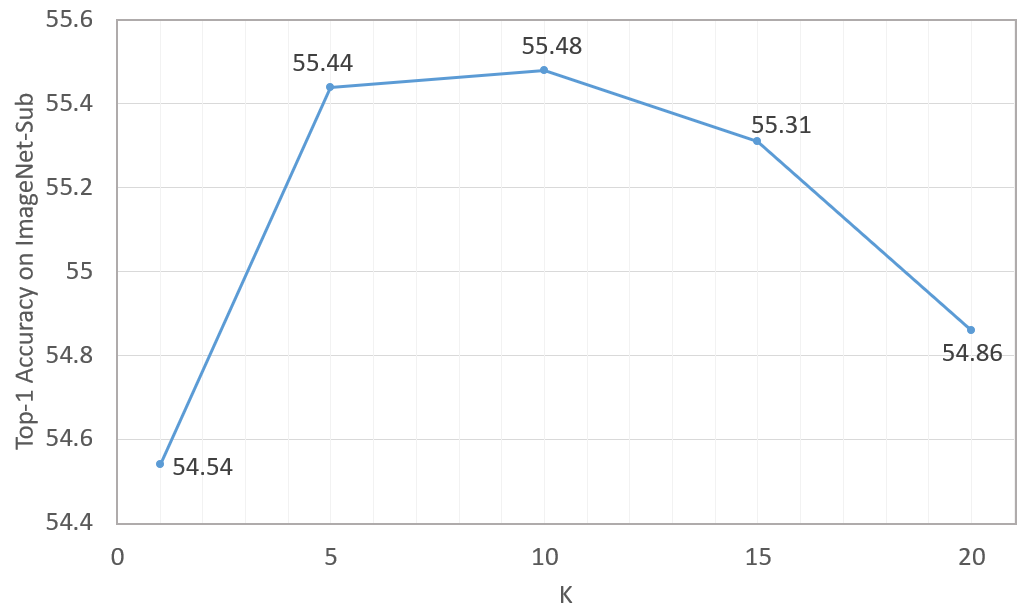}
	\caption{Linear evaluation top-1 accuracy of different value of K on ImageNet-Sub under 10 incremental steps setting.}
	\label{fig:K}
\end{figure}

Besides, since it is impossible to decide the category number of the unlabeled data, the hyper-parameter K in the K-Means algorithm in our experiment is the prior knowledge. Fig. \ref{fig:K} shows that the top-1 accuracy on ImageNet-Sub varies according to the choice of the hyper-parameter K. We find when K varies within a certain range, from 5 to 15, the performance of our method has little change. However, when the gap between K and the real number of classes is too large, the accuracy decreases rapidly.

\begin{table}[t]
	\begin{center}
	\caption{\label{tab:size}
		Linear evaluation top-1 accuracy of different size of the extra sample queue. The best result is achieved when the size is 128.
	}
		\begin{tabular}{|c|c|}
			\hline  
			Size & Top-1 Acc\\
			\hline  
			32 & 53.01~$\pm$~0.33\\
			64 &53.61~$\pm$~0.28\\
			\textbf{128}&\textbf{53.80}~$\pm$~0.37\\
			256&53.59~$\pm$~0.35\\
			512&53.23~$\pm$~0.34\\
			\hline  
		\end{tabular}
	\end{center}
\end{table}
We further explore the effect of the size of the extra sample queue and conduct the experiment on CIFAR-100. Table \ref{tab:size} shows that when the size of the extra sample queue increases, the performance is not improved consistently. A possible reason is that, when the size is large, the update rate of the extra sample queue is slow. Thus, many feature vectors in the extra sample queue are from previous iterations, which are detrimental to the update of the model.\par

\subsection{Limitations}
In this paper, we try to solve the problem of continual contrastive learning. The experiment results show that our method works well on both small and large datasets. However, our work still has many limitations. First, our method is only available for the contrastive learning method. When applying our method to other self-supervised learning methods, like GAN~\cite{gan}, BYOL~\cite{byol}, and MAE~\cite{mae}, some additional revisions are required.
Second, though our method narrows the gap between continual contrastive learning and its upper bound, there is still a large gap between them. We hope our work can give researchers some insights into the catastrophic forgetting problems in contrastive learning, and inspire more research in this direction.

\section{Conclusion}
In this paper, we propose a rehearsal-based continual contrastive learning framework, to alleviate the catastrophic forgetting in contrastive learning. Our method restores a small number of images of old data with a novel sampling strategy and rehearses them while learning the new dataset. Besides, we exploit self-supervised knowledge distillation and propose an extra sample queue to make the network learn a better feature representation from old and new data. Extensive experimental results and analyses demonstrate the effectiveness of our method.

\bibliographystyle{IEEEbib}
\bibliography{icme2022template}

\def\x{{\mathbf x}}
\def\L{{\cal L}}

\renewcommand\thesection{\Alph{section}}
\renewcommand\thefigure{\Alph{figure}}
\renewcommand\thetable{\Alph{table}}
\setcounter{section}{0}
\setcounter{figure}{0}
\setcounter{table}{0}

\section*{Appendix}
\section{Pseudo-code}
We provide a Pseudo-code of our method here.

\begin{algorithm}[!h]
\caption{\label{alg:mmcl} Pseudo-code for training}
    \quad \textbf{Input:} Dataset $\{\bm{D_1},\bm{D_2},...,\bm{D_N}\}$, encoder $\bm{f^S_q}$, $\bm{f^S_k}$ and $\bm{f^T_q}$, augmentation $\bm{\mathcal{T}}$, queue $\bm{Q}$, extra sample queue $\bm{ESQ}$, resotred data set $\bm{S}$ \par
    \quad $\bm{S\leftarrow \emptyset}$\par
    \quad \textbf{for} ${t = 1,...,N}$ \textbf{do} \par
    \quad \quad $\bm{D}=\bm{D_t}\cup \bm{S}$ \par
    \quad \quad Params($\bm{f^T_q}$) $\leftarrow$  Params($\bm{f^S_q}$)\par
    \quad \quad \textbf{for} iteration = $ {1,...,K}$ \textbf{do} \par
    \quad\quad \quad \textbf{for} minibatch $B=B_D\cup B_S \sim D$ \textbf{do} \par
    
    \quad\quad \quad \quad \textit{\textcolor{gray}{\# get positive and negative pairs}} \par
    \quad\quad \quad \quad $\bm{z^S_q}$ = $\bm{f^S_q}(\bm{\mathcal{T}}(B))$ \par
    \quad\quad \quad \quad $\bm{z^S_k}$ = $\bm{f^S_k}(\bm{\mathcal{T}}(B))$ \par
    \quad\quad \quad \quad \textit{\textcolor{gray}{\# calculate MoCo loss and ESQ loss}} \par
    \quad\quad \quad \quad $\bm{\mathcal{L}}_{MoCo}$ = Contrastive$\_$loss($\bm{z^S_q}, \bm{z^S_k}, \bm{Q}$) \par
    \quad\quad \quad \quad $\bm{\mathcal{L}}_{ESQ}$ = Contrastive$\_$loss($\bm{z^S_q}, \bm{z^S_k}, \bm{ESQ}$) \par
    \quad\quad \quad \quad \textit{\textcolor{gray}{\# knowledge distillation}} \par
    \quad\quad \quad \quad \textit{\textcolor{gray}{\# get embeddings}} \par
    \quad\quad \quad \quad $\bm{z^S}$, ~$\bm{z^T}$ = $\bm{f^S_q}(B_S)$, ~$\bm{f^T_q}(B_S)$ \par
    \quad\quad \quad \quad $\bm{z^S_q}$, ~$\bm{z^T_q}$ = $\bm{f^S_q}(\bm{\mathcal{T}}(B_S))$,~ $\bm{f^T_q}(\bm{\mathcal{T}}(B_S))$ \par
    \quad\quad \quad \quad \textit{\textcolor{gray}{\# calculate the similarity matrix}} \par
    \quad\quad \quad \quad $\bm{P^T}$ = Softmax(Similarity($\bm{Z^T}$, $\bm{Z^T_q}$)/$\tau$, dim=-1) \par
    \quad\quad \quad \quad $\bm{P^S}$ = Softmax(Similarity($\bm{Z^S}$, $\bm{Z^S_q}$)/$\tau$, dim=-1) \par
    \quad\quad \quad \quad \textit{\textcolor{gray}{\# calculate knowledge distillation loss}} \par
    \quad\quad \quad \quad $\bm{\mathcal{L}}_{kd}$ = KL$\_$divergence($\bm{P^T}$, $\bm{S^T}$) \par
    \quad\quad \quad \quad $\bm{\mathcal{L}}_{total}$ = $\lambda_1 \bm{\mathcal{L}}_{MoCo}$ + $\lambda_2 \bm{\mathcal{L}}_{ESQ}$ + $\lambda_3\bm{\mathcal{L}}_{kd}$ \par
    \quad\quad \quad \quad \textit{\textcolor{gray}{\# update the model}} \par
    \quad\quad \quad \quad Update($\bm{f^S_q}$, $\bm{\mathcal{L}}_{total}$) \par
    \quad\quad \quad \quad \textit{\textcolor{gray}{\# update the momentum encoder}} \par
    \quad\quad \quad \quad Momentum$\_$update($\bm{f^S_q}$, $\bm{f^S_k}$, $m$) \par
    \quad\quad \quad \quad \textit{\textcolor{gray}{\# update the queue and the extra sample queue}} \par
    \quad\quad \quad \quad Queue\_update($\bm{Q}$, $\bm{z^S_k}$) \par
    \quad\quad \quad \quad $\bm{k_S}$ = $\bm{f^S_k}(\bm{\mathcal{T}}(B_S))$ \par
    \quad\quad \quad \quad Queue\_update($\bm{ESQ}$, $\bm{k_S}$) \par
    
    \quad\quad \quad \textbf{end for} \par
    \quad\quad \quad \textit{\textcolor{gray}{\# update the teacher network}} \par
    \quad\quad \quad Momentum$\_$update($\bm{f^S_q}$, $\bm{f^T_q}$, $m_t$) \par
    \quad \quad \textbf{end for} \par
    \quad\quad \textit{\textcolor{gray}{\# get samples}} \par
    \quad \quad $\bm{S}$ = $\bm{S}\cup$ Sample$\_$strategy($\bm{D_t}$)
    
        
    \quad \textbf{end for}\par
    \quad \textbf{return} $\bm{f^S_q}$
    
\end{algorithm}












    

\section{Additional results}
Besides linear evaluation, we also evaluate our method under \textit{Forgetting} and \textit{Forward Transfer} metrics, that is,
\begin{equation*}
    Forgetting:~ F=\frac{1}{T-1}\sum_{i=1}^{T-1}\max_{t\in \{1,...,T\}}(a_{t,i}-a_{T,i}),
\end{equation*}
\begin{equation*}
    Forward~Transfer:~ FT=\frac{1}{T-1}\sum_{i=2}^{T}(a_{i-1,i}-R_i),
\end{equation*}
where $a_{i,j}$ is the linear evaluation accuracy of the model on dataset $D_j$ after observing the last sample from dataset $D_i$, and $R_i$ is the linear evaluation accuracy of the model with random initialization on dataset $D_i$.\par
The results are shown in the Table \ref{fig:tft}. We can find that our method outperforms \textit{Finetuning} under both \textit{Forgetting} and \textit{Forward Transfer} evaluation metrics. The results further demonstrate that our method can alleviate the catastrophic forgetting in self-supervised learning.
\begin{table}[!h]
	\begin{center}
	\caption{
 \textit{Forgetting} and \textit{Forward Transfer} results on ImageNet-Sub under 5/10 incremental steps settings.}
	\label{fig:tft}
	\begin{tabular}{|c|c|c|c|c|}
		\hline  
		\multirow{2}{*}{Method} &\multicolumn{2}{c|}{\textbf{T=5}}&\multicolumn{2}{c|}{\textbf{T=10}}\\ 

    \cline{2-5}
    & F($\downarrow$) & TF($\uparrow$) & F($\downarrow$) & TF($\uparrow$) \\
		 \hline
		 Finetuning & 0.7&47.1&3.5&45.8\\
		 Simple Rehearsal & \textbf{0.3}&\textbf{48.4}&2.0&47.0 \\
		 Our & \textbf{0.3} &48.3&\textbf{1.6}&\textbf{47.3}\\
		\hline  
		\end{tabular}
	\end{center}
\end{table}

\section{Implementation Detail}
\label{id}
To demonstrate the generalization of our method, we apply our method to SimCLR and InsDisc. We give the implementation detail here. \par 
There are two loss terms, $\mathcal{L}_{kd}$ and $\mathcal{L}_{ESQ}$, should be added. First, the knowledge distillation term can be directly added to SimCLR and InsDisc without any modification. Second, the extra sample queue provides the extra negative samples from old data. For both SimCLR and InsDisc, these extra negative samples are used to calculate contrastive loss with their original positive samples, and form the ESQ loss term.\par
The hyperparameters, including balancing weights, temperature and the size of the extra sample queue, are the same as those in the MoCoV2.

\end{document}